\documentclass[runningheads]{llncs}
\usepackage[T1]{fontenc}
\usepackage{graphicx}

\usepackage[hidelinks]{hyperref}
\usepackage{booktabs}
\usepackage{algorithm}
\usepackage{algorithmic}
\usepackage{mathtools}
\usepackage{amssymb}
\usepackage{tabularx}
\usepackage{multirow}

\usepackage{caption}
\usepackage{subcaption}
\usepackage{comment}
\usepackage{graphicx}
\usepackage{csquotes}
\usepackage{enumitem}

\usepackage{color}

\newcommand*{\CAV}{\ensuremath{\text{CAV}}}
\newcommand*{\attr}{\ensuremath{\text{attr}}}

\begin{document}
\title{Evaluating the Stability of Semantic Concept Representations in CNNs for Robust Explainability}
%
\titlerunning{Evaluating the Stability of Semantic Concept Representations in CNNs}
%

\author{
Georgii Mikriukov\inst{1,2}\orcidID{0000-0002-2494-6285}
\and Gesina Schwalbe\inst{1}\orcidID{0000-0003-2690-2478}
\and Christian Hellert\inst{1}\orcidID{0000-0002-5781-6575}
\and Korinna Bade\inst{2}\orcidID{0000-0001-9139-8947}
}
\authorrunning{G. Mikriukov et al.}
%
\institute{
Continental AG, Germany\\
\email{\{firstname.lastname\}@continental-corporation.com}\\
\and
Hochschule Anhalt, Germany\\
\email{\{firstname.lastname\}@hs-anhalt.de}
}

\maketitle

\begin{abstract}

Analysis of how semantic concepts are represented within Convolutional Neural Networks (CNNs) is a widely used approach in Explainable Artificial Intelligence (XAI) for interpreting CNNs. A motivation is the need for transparency in safety-critical AI-based systems, as mandated in various domains like automated driving.
However, to use the concept representations for safety-relevant purposes, like inspection or error retrieval, these must be of high quality and, in particular, stable.
This paper focuses on two stability goals when working with concept representations in computer vision CNNs: stability of concept retrieval and of concept attribution. The guiding use-case is a post-hoc explainability framework for object detection (OD) CNNs, towards which existing concept analysis (CA) methods are successfully adapted.
To address concept retrieval stability, we propose a novel metric that considers both concept separation and consistency, and is agnostic to layer and concept representation dimensionality. We then investigate impacts of concept abstraction level, number of concept training samples, CNN size, and concept representation dimensionality on stability.
For concept attribution stability we explore the effect of gradient instability on gradient-based explainability methods.
The results on various CNNs for classification and object detection yield the main findings that (1) the stability of concept retrieval can be enhanced through dimensionality reduction via data aggregation, and (2) in shallow layers where gradient instability is more pronounced, gradient smoothing techniques are advised.
Finally, our approach provides valuable insights into selecting the appropriate layer and concept representation dimensionality, paving the way towards CA in safety-critical XAI applications.


\end{abstract}

\section{Introduction}
\label{sec:intro}

Advancements in deep learning in the last decade have led to the ubiquitous use of deep neural networks (DNNs), in particular CNNs, in computer vision (CV) applications like object detection. While they exhibit state-of-the-art performance in many fields, their decision-making logic stays opaque and unclear due to their black-box nature~\cite{bodria2021benchmarking,vilone2021classification}. This fact raises concerns about their safety and fairness, which are desirable in fields like automated driving or medicine. These demands are formalized in industrial standards or legal regulations. For example, the ISO26262~\cite{iso26262} automotive functional safety standard recommends manual inspectability, and the General Data Protection Regulation~\cite{goodman2017european} as well as the upcoming European Union Artificial Intelligence Act~\cite{veale2021demystifying} both demand algorithm transparency. The aforementioned concerns are subject of XAI.

XAI is a subfield of AI that focuses on revealing the inner workings of black-box models in a way that humans can understand~\cite{schwalbe_comprehensive_2023,carvalho_machine_2019,linardatosExplainableAIReview2021}. One approach involves associating semantic concepts from natural language with internal representations in the DNN's latent space~\cite{schwalbe_comprehensive_2023}. In computer vision, a semantic concept refers to an attribute that can describe an image or image region in natural language (e.g., \enquote{pedestrian head}, \enquote{green})~\cite{fong2018net2vec,kim2018interpretability}. These concepts can be associated with vectors in the CNN's latent space, also known as concept activation vectors (CAVs)~\cite{kim2018interpretability}. Post-hoc CA involves acquiring and processing CAVs from trained CNNs~\cite{kim2018interpretability,pfau_robust_2021,abid_meaningfully_2022}, which can be used to quantify how concepts attribute to CNN outputs and apply it to verification of safety~\cite{schwalbe2021verification} or fairness~\cite{kim2018interpretability}.
%
However, in literature two paradigms of post-hoc CA have so far been considered separately, even though they need to be combined to fully compare CNN learned concepts against prior human knowledge. These paradigms are: supervised CA, which investigates pre-defined concept representations~\cite{kim2018interpretability,fong2018net2vec,schwalbe2021verification}, and unsupervised CA, which retrieves learned concepts~\cite{zhang2021invertible,ge2021peek} and avoids expensive labeling costs.
Furthermore, current XAI approaches are primarily designed and evaluated for small classification and regression tasks~\cite{schwalbe2021verification,abid_meaningfully_2022}, whereas more complex object detectors as used in automated driving require scalable XAI methods that can explain specific detections instead of just a single classification output.

Besides adaptation to object detection use-cases, high-stakes applications like safety-critical perception have high demands regarding the quality and reliability of verification tooling~\cite[Chap.\,11]{iso/tc22/sc32_iso_2018h}.
A particular problem is stability: One should obtain similar concept representations given the same CNN, provided concept definitions, and probing data. Instable representations that vary strongly with factors like CA initialization weights \cite{rabold_expressive_2020} or imperceptible changes of the input~\cite{smilkov2017smoothgrad} must be identified and only very cautiously used.
Stability issues may arise both in the retrieval of the concept representations, as well as in their usage. Retrieval instability was already identified as an issue in the base work \cite{kim2018interpretability}, and may lead to concept representations of different quality or even different semantic meaning for the same concept.
Instability in usage may especially occur when determining local concept-to-output attribution. In particular, the baseline approach proposed by Kim et al.~\cite{kim2018interpretability} uses sensitivity, which is known to be brittle with respect to slight changes in the input \cite{sundararajan_axiomatic_2017,smilkov2017smoothgrad}.

This work tackles the aforementioned problems of OD-ready supervised and unsupervised CA, and measurement and improvement of stability in CA retrieval and attribution.
Concretely, to solve these problems, we propose an XAI framework based on supervised and unsupervised CA methods for ODs. The unsupervised method is used to automatically mine concept samples, which are jointly used for supervised concept analysis with manually labeled concepts. Furthermore, stability metrics are suggested and tested. The respective main contributions of our work are:
\begin{itemize}
    \item Proposal of two metrics and methodology for testing of \emph{concept retrieval stability} and \emph{concept attribution stability} in CA;
    \item Experimental study of \emph{stability influence factors} in six diverse CNN models with different backbones with the main findings that \emph{CAV dimensionality reduction may improve stability}, and that \emph{gradient smoothing may be beneficial} for concept attribution stability in shallow layers;
    \item Adaptation of supervised and unsupervised concept-based analysis methods for \emph{CA on common ODs};
    \item Introduction of a post-hoc, label-efficient, concept-based \emph{explainability framework} for classifiers and ODs allowing for concept stability estimation (Fig.\,\ref{fig:main-diagram}).
\end{itemize}

In the following, we will first take a look at related work on concept analysis in Sec.\,\ref{sec:related}. Our approaches for combining supervised and unsupervised CA, for CA in OD, and for stability measurement are then detailed in Sec.\,\ref{sec:method}. Our experimental setup can be found in Sec.\,\ref{sec:setup} with results detailed in Sec.\,\ref{sec:experiments}.

\section{Related Work}
\label{sec:related}


\subsection{Supervised Concept Analysis}
\label{sec:related-supervised}

There are two primary paradigms in supervised CA methods: scalar-concept representation~\cite{koh2020concept,sawada2022concept,chen2020concept} and vector-concept representation~\cite{bau2017network,fong2018net2vec,kim2018interpretability}.
Scalar concept representations refer to disentangled deep neural network (DNN) layer representations with a one-to-one correspondence between neurons and distinct semantic concepts.
A prominent example and base work are Concept Bottleneck Models~\cite{koh2020concept} (CBM). These introduce an interpretable bottleneck layer to DNNs by assigning each neuron to a specific concept, i.e., scalar-concepts. An extension CBM-AUC~\cite{sawada2022concept}, enhances the model's capability by automatically learning unsupervised concepts (AUC) that describe the residual variance of the feature space.
In contrast to the previous examples, Concept Whitening~\cite{chen2020concept} is a post-hoc approach towards scalar-concepts. It transforms a feature space of a layer and reduces redundancy between neurons, making it more likely for each neuron to correspond to a single concept.
IIN~\cite{esser_disentangling_2020} is another post-hoc approach that trains an invertible neural network to map a layer output to a disentangled version, using pairwise labels.
However, standard CNNs are typically highly entangled \cite{kazhdan_disentanglement_2021}. Hence, such scalar-concept approaches have to enforce the disentangled structure during training or utilize potentially non-faithful proxies~\cite{margeloiu2021concept}. Furthermore, they are limited to explaining a single layer.

Vector-concepts, on the other hand, associate a concept with a vector in the latent space.
The base work in this direction still disregarded the distributed nature of CNN representations: The Network Dissection approach~\cite{bau2017network} aims to associate each convolutional filter in a CNN with a semantic concept. Its successor Net2Vec~\cite{fong2018net2vec} corrects this issue by associating a concept with a linear combination of filters, resulting in a concept being globally represented by a vector in the feature space, the concept activation vector (CAV) \cite{kim2018interpretability}. A sibling state-of-the-art method for associating concepts with latent space vectors is TCAV~\cite{kim2018interpretability}, which also uses a linear model attached to a CNN layer to distinguish between neurons (in contrast to filters as in Net2Vec) relevant to a given concept and the rest. TCAV also proposes a gradient-based approach that allows for the evaluation of how sensitive a single prediction or complete class is to a concept. The concept sensitivity (attribution) for a model prediction is calculated by taking the dot product between the concept activation vector and the gradient vector backpropagated for the desired prediction.
These vector-concept baselines for classification (TCAV) and segmentation (Net2Vec) of concepts have been extended heavily over the years, amongst others towards regression concepts \cite{graziani_concept_2020,graziani_regression_2018}, multi-class concepts \cite{kazhdan_now_2020}, and locally linear \cite{wu_global_2020,zhang_examining_2018} and non-linear \cite{kazhdan_now_2020} CAV retrieval. However, the core idea remained untouched.

While the TCAV paper already identifies stability as a potential issue, they reside to significance tests for large series of experiments leaving a thorough analysis of stability (both for concept retrieval and concept attribution) open, as well as investigation of improvement measures. Successor works tried to stabilize the concept attribution measurement. For example, Pfau et al.~\cite{pfau_robust_2021} do not use the gradient directly, but the average change of the output when perturbing the intermediate output towards the CAV direction in latent space in different degrees. This gradient stabilization approach follows the idea of Integrated Gradients \cite{sundararajan_axiomatic_2017}, but no other approaches like Smoothed Gradients \cite{smilkov2017smoothgrad} have been tried.
Other approaches also suggest improved metrics for global concept attribution \cite{graziani_regression_2018}. However,
to our knowledge,
stability remained unexplored so far.

We address this gap by utilizing TCAV as a baseline global concept vector representation for the stability estimation. Moreover, as gradient-based method, it be adapted to estimate concept attributions in other model types, such as ODs (see Sec.~\ref{sec:method-adaptation}). It is important to note that our stability assessment method is not limited to TCAV and can potentially be applied to evaluate the stability of other global concept representations.

\subsection{Unsupervised Concept Analysis}
\label{sec:related-unsupervised}

Unsupervised methods for analyzing concepts are also referred to as concept mining~\cite{schwalbe_concept_2022}. These methods do not rely on pre-defined concept labels, but the acquired concepts are not always meaningful and require manual revision. There are two main approaches to concept mining: clustering and dimensionality reduction. Clustering methods, such as ACE~\cite{ghorbani2019towards} and VRX~\cite{ge2021peek} group latent space representations of image patches (superpixels), obtained through segmentation algorithms. The resulting clusters are treated as separate concepts and can be used for supervised concept analysis. Invertible Concept Extraction (ICE)~\cite{zhang2021invertible} is a dimensionality reduction method based on non-negative matrix factorization. It mines non-negative concept activation vectors (NCAVs) corresponding to the most common patterns from sample activations in intermediate layers of a CNN. The resulting NCAVs are used to map sample activations to concept saliency maps, which show concept-related regions in the input space.

To reduce the need in concept labeling, we opted to use ICE for unsupervised concept mining due to (1) its superior performance regarding interpretability and completeness of mined concepts compared to clustering~\cite{zhang2021invertible}, and (2) its simpler and more straightforward pipeline with less hyperparameters. Unlike ACE, it does not rely on segmentation and clustering results as an intermediate step, which makes it easier to apply.

\subsection{Concept Analysis in Object Detection}
\label{sec:related-ca-od}

There are only a few existing works that apply concept analysis methods to object detection, due to scalability issues. In \cite{schwalbe2021verification} the authors adapt Net2Vec for scalability to OD activation map sizes, which is later used to verify compliance of the CNN behavior with fuzzy logical constraints \cite{schwalbe2022concept}. Other TCAV-based works apply lossy average pooling to allow large CAV sizes \cite{graziani_concept_2020,chyung_extracting_2019}, but do not test OD CNNs.
However, these methods are fully supervised and require expensive concept segmentation maps for training, resulting in scalability issues regarding concept label needs.
In order to reduce the need for concept labels, we propose adapting and using a jointly supervised and unsupervised classification approach for object detection, and investigate the impact of CAV size on stability.
This also closes the gap that,
to our knowledge,
no unsupervised CA method has been applied to OD-sized CNNs so far.

\begin{figure*}[t]
  \centering
  \includegraphics[width=1.0\linewidth]{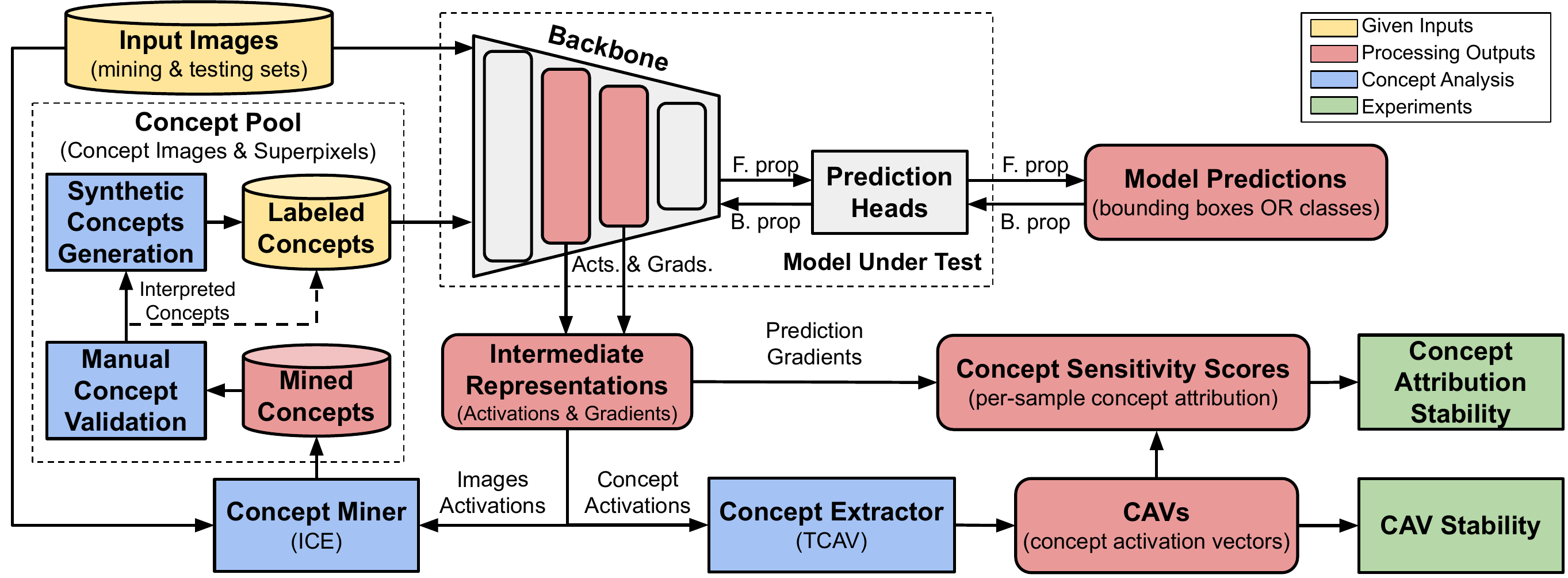}
  \caption{The framework for estimation of CAV stability and concept attribution stability. The proposed solution utilizes unsupervised ICE to aid concept discovery and labeling, while supervised TCAV is used for the generation of concept representations.}
  \label{fig:main-diagram}
\end{figure*}

\section{Proposed Method}
\label{sec:method}

The overall goal targeted here is a CA framework that allows stable, label-efficient retrieval and usage of interpretable concepts for explainability of both classification and OD backbones.
To address this, we introduce a framework that combines unsupervised CA (for semi-automated enrichment of the available concept pool) with supervised CA (for retrieval of CAVs and CNN evaluation) together with an assessment strategy for its stability properties.
An overview of the framework is given in the following in Sec.\,\ref{sec:method-framework}, with details on how we adapted CA for OD in Sec.\,\ref{sec:method-adaptation}.
Sec.\,\ref{sec:method-metrics} then presents our proposal of CAV stability metrics.
Lastly, one of the potential influence factors on stability, namely CAV dimensionality and parameter reduction techniques, is presented in Sec.\,\ref{sec:method-cav-dim}.


\subsection{Stability Evaluation Framework}
\label{sec:method-framework}


The framework depicted in Fig.\,\ref{fig:main-diagram} aims to efficiently combine supervised and unsupervised CA methods for use in explainability or evaluation purposes, like our CA stability evaluation. To achieve this it (1) builds an extensible \textit{Concept Pool} containing human-validated \textit{Mined Concepts} extracted from trained \textit{Model Under Test}, and (optionally) existing manually \textit{Labeled Concepts}; and it (2) uses these concepts to obtain \textit{CAVs} and, e.g., conduct \textit{CAV Stability} and \textit{Concept Attribution Stability} tests on object detection and classification models.

\medskip\noindent\textbf{Concept Pool Creation/Extension.}
In some CV domains, it can be challenging to find publicly available datasets with high-quality concept labels. In order to streamline the manual annotation process and speed up concept labeling, we utilize unsupervised concept mining.
The left side of Fig.\,\ref{fig:main-diagram} depicts the process of creating the \textit{Concept Pool} (or extending it, if we already have an initial set of \textit{Labeled Concepts}) by employing the \textit{Concept Miner}. A concept in the concept pool is represented by a set of images or image patches showing the concept. To extract additional \textit{Mined Concepts}, the \textit{Concept Miner} identifies image patches that cause common patterns in the CNN \textit{Image Activations}. The activations are extracted from the layer of interest of the \textit{Backbone} of the \textit{Model Under Test} for \textit{Input Images} from the mining set.
In our work, we utilize ICE~\cite{zhang2021invertible} as the \textit{Concept Miner} to obtain the image patches. The workflow of ICE is as follows: (1) it first mines NCAVs; then, for each NCAV and each sample from a test set (2) it applies NCAV inference, i.e., obtains a (non-binary) heatmap of where the NCAV activates in the image, and (3) masks the input image with the binarized heatmap. For details see Sec.\ref{sec:related-unsupervised} and \cite{zhang2021invertible}.
The sets of mined image patches, alias concepts, next undergo \textit{Manual Concept Validation}: A human annotator assigns a label to each \textit{Mined Concepts}. These \textit{Interpreted Concepts}, if meaningful, can either directly be added to the set of \textit{Labeled Concepts} or be utilized in \textit{Synthetic Concept Generation} to obtain more complex synthetic concept samples (see Sec.\,\ref{sec:setup-concepts} and Fig.\,\ref{fig:synth-concepts} for more details and visual examples).
It should be noted that the \textit{Concept Pool}, once established, is model-agnostic and can be reused for other models, and that the ICE concept mining approach can be exchanged by any other suitable unsupervised CA method that produces concept heatmaps during inference.

\medskip\noindent\textbf{Concept Stability Analysis.}
Now that the \textit{Concept Pool} is established, we can perform supervised CA to obtain \textit{CAVs} for the concepts in the pool. The CAV training is done on the \textit{Concept Activations}, i.e., CNN activations of concept images from the \textit{Labeled Concepts} in the \textit{Concept Pool}. Given \textit{CAVs}, we can then calculate per-sample concept attribution using, e.g., backpropagation-based sensitivity methods \cite{kim2018interpretability}. The resulting \textit{CAVs} and \textit{Concept Sensitivity Scores} can then be used for local and global explanation purposes. To ensure their quality, this work investigates stability (\textit{CAV Stability} and \textit{Concept Attribution Stability}) of these for OD use-cases, as detailed in Sec.\,\ref{sec:method-metrics}.


For supervised CA we use the base TCAV~\cite{kim2018interpretability} approach: A binary linear classifier
is trained to predict presence of a concept from the intermediate neuron activations in the selected CNN layer. The classifier weights serve as CAV, namely the vector that points into the direction of the concept in the latent space. The CAVs are trained in a one-against-all manner on the labeled concept examples from the \textit{Concept Pool}.
For concept attribution, we adopt the sensitivity score calculation from \cite{kim2018interpretability}:  for a sample is the partial derivative of the CNN output in the direction of the concept, which is calculated as the dot product between the CAV and the gradient vector in the CAV layer.
In this paper, we are interested in the stability of this retrieval process for obtaining CAVs and respective concept attributions.


\subsection{Concept Analysis in Object Detectors}
\label{sec:method-adaptation}

The post-hoc concept stability assessment framework described above, in particular the used TCAV and ICE methods, is out-of-the-box suitable for use with classification models. However, object detection networks pose additional challenges: besides larger sizes, they have different prediction heads and employ suppressive post-processing of the output.

\noindent\textbf{Multiple Predictions.}
Unlike classification models that produce a single set of predictions per sample, object detectors may produce multiple predictions, requiring adaptions to TCAV and ICE.

For ICE the concept weights and importance estimation component require adjustments. The pipeline assesses the effect of small modifications to each concept on the final class prediction. For classification, this estimation is performed on a per-sample basis. For object detection, we switch that calculation to the per-bounding box approach.

The TCAV process of calculating CAVs remains unchanged. However, TCAV employs gradients backpropagated from the corresponding class neuron and concept CAV to assess the concept sensitivity of the desired output class. In object detectors, concept sensitivity can be computed for each prediction, or bounding box, by starting the backpropagation from the desired class neuron of the bounding box.

It is important to note that some object detection architectures predict an objectness score for each bounding box, which can serve as an alternative starting neuron for the backpropagation~\cite{kirchknopf_armin_explaining_nodate}. Nonetheless, we only use class neurons for this purpose in our experiments.

\noindent\textbf{Suppressive Post-processing.}
Another challenge in object detection is explanation of False Negatives (FNs), which refer to the absence of detection for a desired object.
Users may be especially interested in explanations regarding FN areas, e.g., for debugging purposes. While the raw OD CNN bounding box predictions usually cover all image areas, post-processing may filter out bounding boxes due to low prediction certainty or suppress them during Non-Maximum Suppression (NMS). To still evaluate concept sensitivity for FNs, we compare the list of raw unprocessed bounding boxes with the desired object bounding boxes specified by the user. We then use Intersection over Union (IoU) to select the best raw bounding boxes that match the desired ones, and these selected bounding boxes (i.e., their output neurons) are used for further evaluation.


\subsection{Evaluation of Concept Stability}
\label{sec:method-metrics}

\medskip\noindent\textbf{Concept Retrieval Stability.}
We are interested in concepts that are both \emph{consistent} and \emph{separable} in the latent space. However, these two traits have not been considered jointly in previous work. Thus, we define the generalized concept stability $\mathcal{S}_{L_{k}}$ metric for a concept $C$ in layer $L_{k}$ applicable to a test set $X$ as
\begin{align}
    \mathcal{S}_{L_{k}}^C(X) \coloneqq \texttt{separability}_{L_{k}}^C(X) \times \texttt{consistency}_{L_{k}}^C,
\label{eq:concept-stability}
\end{align}

where, $\texttt{separability}_{L_{k}}^C(X)$ represents how well tested concepts are separated from each other in the feature space, $\texttt{consistency}_{L_{k}}^C$ denotes how similar are representations for the same concept when obtained with different initialization conditions.

\textit{Separability.}
The binary classification performance of each CAV reflects how effectively the concept is separated from other concepts, when evaluated in a concept-vs-other manner rather than a concept-vs-random approach. In the concept-vs-other scenario, the non-concept-class consists of all other concepts, whereas it is a single randomly selected other concept in the concept-vs-random scenario~\cite{kim2018interpretability}.
We choose the separability from Equ.\,\ref{eq:concept-stability} for a single concept $C$ on the test set $X$ as:
\begin{align}
    \textstyle
    \texttt{separability}_{L_{k}}^C(X)
    \coloneqq f1_{L_{k}}^C(X)
    \coloneqq \frac{1}{N} \sum_{i=1}^N f1(CAV_{L_k,i}^C; X)
    \in [0,1]
\label{eq:cav-stability}
\end{align}
where $f1_{L_{k}}^C$ is the mean of relative F1-scores $f1(-;X)$ on $X$ for $\CAV_{L_k,i}^C$ of $C$ in layer $L_k$ for $N$ runs $i$ with different initialization conditions for CAV training.

\textit{Consistency.}
In TCAV, during the CAVs training, a limited amount of concept samples may lead to model underfitting, and significant inconsistency between CAVs obtained for different training samples and initialization conditions~\cite{kim2018interpretability}. Since cosine similarity was shown to be a suitable similarity measure for CAVs \cite{kim2018interpretability,fong2018net2vec} we set the consistency measure to the mean cosine similarity between the CAVs in layer $L_k$ of $N$ runs:
\begin{align}
    \SwapAboveDisplaySkip
    \texttt{consistency}_{L_{k}}^C
    \coloneqq \cos_{L_{k}}^C
    \coloneqq \tfrac{2}{N(N-1)}\sum_{i=1}^{N}
    \sum_{j=1}^{i-1} {\cos(\CAV_{L_k,i}^C, \CAV_{L_k,j}^C)}
    \,,
    \label{eq:cav-stability-cos}
\end{align}
where $\cos(-,-)$ is cosine similarity, here between CAVs of the same concept $C$ and layer $L_k$ obtained during different runs $i,j$.

\medskip\noindent\textbf{Concept Attribution Stability.}
Small changes in the input space may significantly change the output and, thus, the gradient values. TCAV requires gradients to calculate the concept sensitivity (attribution) of given prediction. Hence, gradient instability may have an impact on the explanations, and, in the worst case, change it from positive to negative attribution or vice versa.

We want to check, if such instability of gradient values influences concept detection.
For this, we compare the vanilla gradient approach against a stabilized version using the state-of-the-art gradient stabilization approach
SmoothGrad~\cite{smilkov2017smoothgrad}.
It diminishes or negates the gradient instability in neural networks by averaging vanilla gradients obtained for multiple copies of the original sample augmented with a minor random noise.
For comparison purposes, first the vanilla gradient is propagated backward with respect to the detected object's class neuron. This neuron is remembered and used then for the gradient backpropagation for noisy copies of SmoothGrad.
TCAV concept attributions can naturally be generalized to Smoothgrad, defining them as:
\begin{align}
\label{eq:grad-stability-attr}
    \attr_{C}^{*}(x) \coloneqq \CAV_{C} \circ \nabla^{*} f_{L_k\to}(f_{\to L_k}(x))
    \;,
\end{align}
where $\attr_{C}^{*}$ is the attribution of concept $C$ in layer $L_k$ for vanilla gradient ($*=\text{grad}$) or SmoothGrad ($*=\text{SG}$) for a single prediction for sample $x$, $\CAV_C=\CAV_{L_k,.}^C$, and $f_{\to L_k}$ is the CNN part up to $L_k$, $f_{L_k\to}$ the mapping from $L_k$ representations to the score of the selected prediction and class.

\textit{Acc.}
As one approach, for each tested layer we build a confusion matrix for multiple test samples and bounding boxes therein, where $y_\text{true} = \text{sign}(\attr_{i}^\text{grad})$ and $y_\text{predicted} = \text{sign}(\attr_{i}^\text{SG})$ are predictions to compare the sign of concept attribution for SmoothGrad and vanilla gradient. On this, accuracy (Acc) is used to show the fraction of cases where SmoothGrad and vanilla gradient concept attributions have the same sign, i.e., where gradient instability has no impact.

\textit{CAD.}
As a second approach, to qualitatively evaluate the difference between the concept attribution of SmoothGrad and the vanilla gradient in the tested layer, we introduce the Concept Attribution Deviation (CAD) metric. It shows the average absolute attribution value change for all used concepts $C$ and $N$ runs, and, thus, describes the impact of gradient instability on concept attribution in a layer:
\begin{align}
    \SwapAboveDisplaySkip
    \text{CAD}(x) \coloneqq \frac{
        \textstyle\sum_{C}\sum_{i}^{N} \left |\attr_{C,i}^\text{grad}(x) - \attr_{C,i}^\text{SG}(x)  \right |}{
        \textstyle\sum_{C}\sum_{i}^{N} \left |\attr_{C,i}^\text{grad}(x) \right |}
    \;.
    \label{eq:grad-stability-cad}
\end{align}


\begin{figure*}[t]
  \centering
  \includegraphics[width=0.8\linewidth]{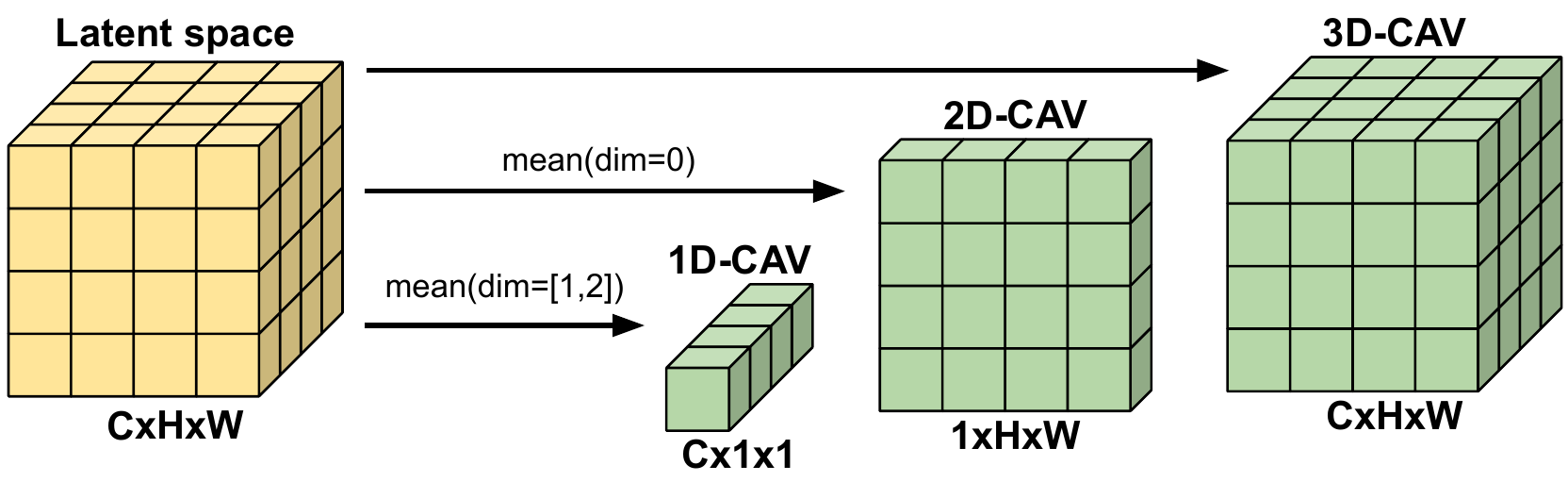}
  \caption{Concept activation vectors (CAVs) of different dimensions.}
  \label{fig:cav-dims}
\end{figure*}

\subsection{CAV Dimensionality}
\label{sec:method-cav-dim}

The stability can be greatly affected by the number of CAV parameters, which is especially important in object detectors with large intermediate representations. Moreover, the larger CAV size leads to increased memory and computation requirements. The original TCAV paper proposes using 3D-CAV-vectors~\cite{kim2018interpretability}. However, alternative translation invariant 1D-~\cite{fong2018net2vec,zhang2021invertible} and channel invariant 2D-CAV-representations, which have less parameters, are possible. If 3D-CAV's dimensions of OD's arbitrary intermediate layer are $C \times H \times W$, then dimensions of 1D- and 2D-CAV are $C \times 1 \times 1$ and $1 \times H \times W$ respectively, where $C$, $H$ and, $W$ denote $channel$, $height$ and, $width$ dimensions respectively (see Fig.\,\ref{fig:cav-dims}).

The 1D-CAV provides during inference one presence score per channel, and possesses the property of translation invariance. This implies that only the presence or absence of a concept in the input space matters, rather than its size or location. In contrast, the 2D-CAV concentrates solely on the location of the concept, providing one presence score for each activation map pixel location. This can also be advantageous in certain circumstances (e.g., for the concepts \enquote{sky} or \enquote{ground}). The 3D-CAV provides during inference a single concept presence score for the complete image, depending both on location, size, and filter distribution of the concept. Meanwhile, it comes with the disadvantage of larger size and higher computational requirements.

Original 3D-CAVs do not require special handling of the latent space. But for evaluation of 1D- and 2D-CAVs, we preprocess incoming latent space vectors to match the CAV dimensionality by taking the mean along width and height, or channel dimensions respectively, as already successfully applied in previous work \cite{chyung_extracting_2019,graziani_concept_2020}. In other words, for the calculation of CAV with reduced dimensions, we aggregate activation functions and gradients along certain dimensions. CAV dimension size is a hyperparameter, which may impact CAV memory consumption, CAV stability, the overall performance of concept separation, CAV training speed, and following operations with CAVs (e.g., evaluation of the concept attribution). Thus, we also propose using our stability metrics for the selection of the optimal CAV dimension size.

\section{Experimental Setup}
\label{sec:setup}

We use the proposed framework to conduct the following experiments for OD and classification models: 1) evaluation of concept representation stability via the selection of representation dimensionality; 2) inspection of the impact of gradient instability in CNNs on concept attribution.
The process of concept analysis in classifiers can be carried out using the default approaches proposed in the original papers~\cite{kim2018interpretability,zhang2021invertible}, and it does not require any special handling.

In the following subsections, we describe selected experimental datasets and concept data preparation, models, model layers, and hyperparameter choices. Experiment results and interpretation are described later in Section~\ref{sec:experiments}.


\subsection{Datasets}
\label{sec:setup-datasets}

\medskip\noindent\textbf{Object Detection.}
For unsupervised concept mining in object detectors and experiments with ODs, we use the validation set of MS COCO 2017~\cite{lin2014microsoft} dataset, containing $5000$ real world images with 2D object bounding box annotations, including many outdoor and urban street scenarios. 
We mine concepts from bounding boxes of \textit{person} class with the area of at least $20000$ pixels, so the mined concept images have reasonable size and can be visually analyzed by a human. The resulting subset includes more than $2679$ bounding boxes of people in different poses and locations extracted from $1685$ images.

\medskip\noindent\textbf{Classification.}
For concept stability experiments with classification model, we use BRODEN~\cite{bau2017network} and CycleGAN Zebras~\cite{zhu2017unpaired} datasets. BRODEN contains more than 60,000 images image and pixel-wise annotations for almost 1200 concepts of 6 categories. CycleGAN Zebras contains almost 1500 images of zebras suitable for supervised concept analysis.



\subsection{Models}
\label{sec:setup-models}

To evaluate the stability of semantic representations in the CNNs of different architectures and generations, we selected three object detectors and three classification models with various backbones.

\medskip\noindent\textbf{Object Detection models:}
    \begin{itemize}[leftmargin=2em, nosep]
        \item one-stage YOLOv5s\footnote[1]{\url{https://github.com/ultralytics/yolov5}}~\cite{glennjocher20204154370} (residual DarkNet~\cite{redmon2018yolov3,he2016deep} backbone);
        \item two-stage FasterRCNN\footnote[2]{\url{https://pytorch.org/vision/stable/models/faster_rcnn}}~\cite{ren2015faster} (inverted residual MobileNetV3~\cite{howard2019searching} backbone);
        \item one-stage SSD\footnote[3]{\url{https://pytorch.org/vision/stable/models/ssd}}~\cite{liu2016ssd} (VGG~\cite{simonyan2014very} backbone).
    \end{itemize}
    
All evaluated object detection models are pre-trained on MS COCO~\cite{lin2014microsoft} dataset. The models are further referred to as YOLO5, RCNN, and SSD.

\medskip\noindent\textbf{Classification models:}
    \begin{itemize}[leftmargin=2em, nosep]
        \item residual ResNet50\footnote[4]{\url{https://pytorch.org/vision/stable/models/resnet}}~\cite{he2016deep};
        \item compressed SqueezeNet1.1\footnote[5]{\url{https://pytorch.org/vision/stable/models/squeezenet}}~\cite{iandola2016squeezenet}
        \item inverted residual EfficientNet-B0\footnote[6]{\url{https://pytorch.org/vision/stable/models/efficientnet}}~\cite{tan2019efficientnet}
    \end{itemize}

Classification models are pre-trained on ImageNet1k~\cite{deng2009imagenet} dataset. The models are further referred to as ResNet, SqueezeNet, and EfficientNet.


\begin{table}

\setlength{\tabcolsep}{2pt}
\centering
\caption{Shorthands $l_i$ of selected classification CNN intermediate layers for Concept Analysis (l=layer, b=block, f=features, squeeze=s).}
\label{tab:layers-cls}
\newcommand{\layerID}[1]{\scriptsize{}#1}
\begin{tabular}{c|c|c|c|c|c|c|c}
\hline
\multirow{2}{*}{Classifier} & \multicolumn{7}{c}{layers}\\ \cline{2-8}
& $l_1$ & $l_2$ & $l_3$ & $l_4$ & $l_5$ & $l_6$ & $l_7$ \\

\hline
\scriptsize ResNet &
\layerID{l1.1.c3} & \layerID{l2.0.c3} & \layerID{l2.2.c3} & \layerID{l3.1.c3} & \layerID{l3.4.c3} & \layerID{l4.0.c3} & \layerID{l4.2.c3} \\

\hline
\scriptsize SqueezeNet &
\layerID{f.3.s} & \layerID{f.4.s} & \layerID{f.6.s} & \layerID{f.7.s} & \layerID{f.9.s} & \layerID{f.10.s} & \layerID{f.11.s}\\

\hline
\scriptsize EfficientNet &
\layerID{f.1.0.b.2.0} & \layerID{f.2.0.b.3.0} & \layerID{f.3.0.b.3.0} & \layerID{f.4.0.b.3.0} & \layerID{f.5.0.b.3.0} & \layerID{f.6.0.b.3.0} & \layerID{f.7.0.b.3.0} \\

\hline
\end{tabular}
\end{table}

\begin{table}

\setlength{\tabcolsep}{2pt}
\centering
\caption{Shorthands $l_i$ of selected OD CNN intermediate layers for Concept Analysis (b=block, f=features, e=extra, c=conv).}
\label{tab:layers-od}
\newcommand{\layerID}[1]{\scriptsize{}#1}
\begin{tabular}{c|c|c|c|c|c|c|c|c|c|c}
\hline
\multirow{2}{*}{OD} & \multicolumn{10}{c}{layers}\\ \cline{2-11}
& $l_1$ & $l_2$ & $l_3$ & $l_4$ & $l_5$ & $l_6$ & $l_7$ & $l_8$ & $l_9$ & $l_{10}$ \\

\hline
\scriptsize YOLO5 &
\layerID{3.c} & \layerID{4.cv3.c} & \layerID{5.c} & \layerID{6.cv3.c} & \layerID{7.c} & \layerID{8.cv3.c} & \layerID{10.c} & \layerID{14.c} & \layerID{17.cv3.c} & \layerID{18.c} \\

\hline
\scriptsize RCNN &
\layerID{3.b.2.0} & \layerID{4.b.3.0} & \layerID{5.b.3.0} & \layerID{6.b.3.0} & \layerID{7.b.2.0} & \layerID{8.b.2.0} & \layerID{10.b.2.0} & \layerID{11.b.3.0} & \layerID{13.b.3.0} & \layerID{15.b.3.0} \\

\hline
\scriptsize SSD &
\layerID{f.5} & \layerID{f.10} & \layerID{f.14} & \layerID{f.17} & \layerID{f.21} & \layerID{e.0.0} & \layerID{e.1.0} & \layerID{e.2.0} & \layerID{e.3.0} & \layerID{e.4.0} \\

\hline
\end{tabular}
\end{table}

\subsection{Layer Selection for Concept Analysis}
\label{sec:setup-layers}

To identify any influence of the layer depth on extracted concept stability, we must analyze the latent space of DNNs across multiple layers. To accomplish this, we extract intermediate representations and concepts from ten intermediate convolutional layers of ODs and seven intermediate convolutional layers of classifiers. These layers are uniformly distributed throughout the backbones of CNNs. The names of the selected layers for each network are listed in Tab.\,\ref{tab:layers-cls} and Tab.\,\ref{tab:layers-od}, where each layer is identified by a symbolic name in the format of $l_x$, where $x$ denotes the relative depth of the layer in the backbone (i.e., layers from $l_1$ to $l_7$ for classifiers and from $l_1$ to $l_{10}$ for ODs).

In experiments, we use semantic concepts of medium-level (e.g., composite shapes) or high-level (e.g., human body parts) abstraction (Sec.\,\ref{sec:setup-concepts}). Shallow layers are ignored, as they mostly recognize concepts of low-level abstraction (e.g., color, texture), whilst deeper layers recognize complex objects and their parts~\cite{wang2020chain,zhang2018interpretable}.


\begin{figure*}[t]
     \centering
     \begin{subfigure}[b]{0.32\textwidth}
         \centering
         \includegraphics[width=\textwidth]{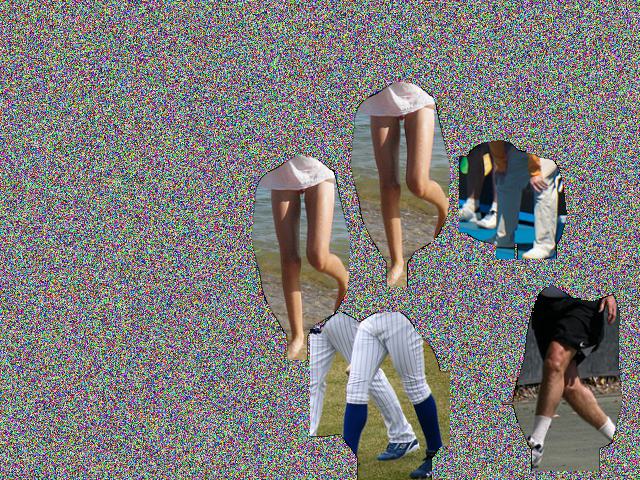}
         \caption{Concept: \enquote{legs}}
         \label{fig:synth-concept-legs}
     \end{subfigure}     
     \hfill     
     \begin{subfigure}[b]{0.32\textwidth}
         \centering
         \includegraphics[width=\textwidth]{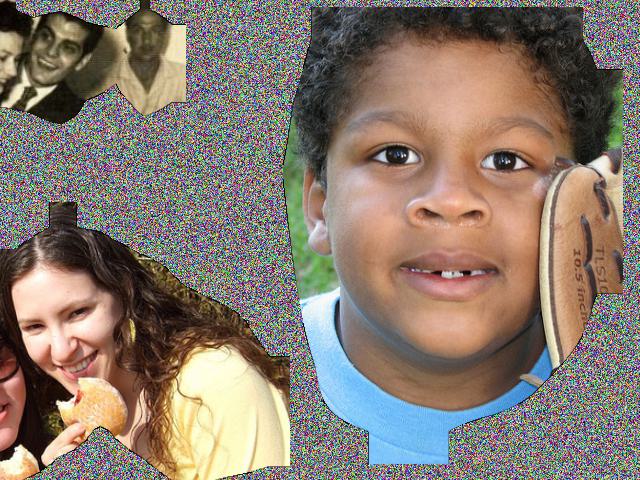}
         \caption{Concept: \enquote{head}}
         \label{fig:synth-concept-head}
     \end{subfigure}     
     \hfill     
     \begin{subfigure}[b]{0.32\textwidth}
         \centering
         \includegraphics[width=\textwidth]{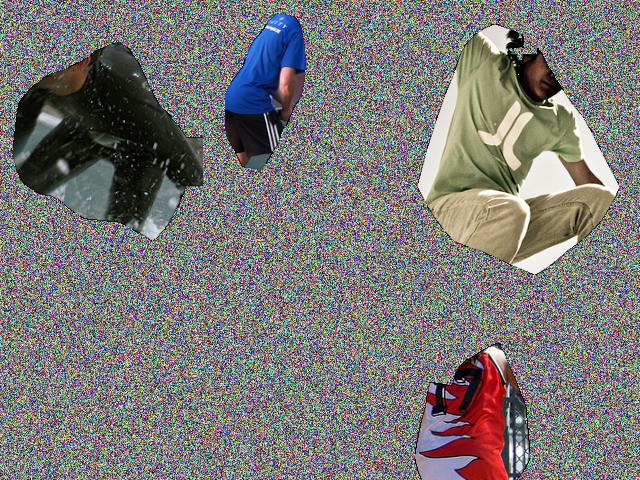}
         \caption{Concept: \enquote{torso}}
         \label{fig:synth-concept-torso}
     \end{subfigure}     
    \caption{Examples of synthetic concept samples generated using concept superpixels obtained from MS~COCO.}
    \label{fig:synth-concepts}
\end{figure*}

\subsection{Synthetic Concept Generation and Concept Selection}
\label{sec:setup-concepts}

\medskip\noindent\textbf{Object Detection.}
To conduct concept analysis experiments with object detectors, we generate synthetic concept samples using concept information extracted from MS COCO (see Fig.\,\ref{fig:main-diagram} and Sec.\,\ref{sec:method-framework}). We used ICE~\cite{zhang2021invertible} to mine concept-related superpixels (image patches) from MS~COCO bounding boxes of the \textit{person} class that have an area of at least 20,000 pixels. Then, we visually inspected 30 mined concepts (10 for each following YOLO5 layer: \texttt{8.cv3.c}, \texttt{9.cv1.c}, and \texttt{10.c}; see caption of Tab.~\ref{tab:layers-od} for notations) and selected 3 concepts semantically corresponding to labels \enquote{legs}, \enquote{head}, and \enquote{torso}.
Interestingly, we found that several concepts (e.g., \enquote{head}, \enquote{legs}) were present in more than one layer. We only picked one of the concepts of the same type based on the subjective quality. 
For each selected concept, we save 100 concept-related superpixels using a concept mask binarization threshold of $0.5$.

Examples of the MS COCO synthetic concepts can be seen in Figure~\ref{fig:synth-concepts}. To generate a synthetic concept sample of a size of $640 \times 480$ pixels, 1 to 5 concept-related superpixels are selected and placed on a background of random noise drawn from a uniform distribution (alternatively, images of natural environments can be used as a background). Additionally, random scaling is applied to the superpixels before placement with a random factor between 0.9 and 1.1.

\medskip\noindent\textbf{Classification.}
We use labeled concepts \enquote{stripes}, \enquote{zigzags}, and \enquote{dots} from BRODEN dataset to analyze the stability of concept representation and attribution in classification models on the examples of zebra images from the CycleGAN dataset.


\subsection{Experiment-specific Settings}
\label{sec:setup-experiments}

\medskip\noindent\textbf{Experiment 1: CAV Stability and Dimensionality.}
We conduct CAV-stability experiments for 1D-, 2D-, and 3D-CAVs (see Sec.~\ref{sec:method-cav-dim}) with YOLO5, RCNN, SSD,  ResNet, SqueezeNet, and EfficientNet models to measure the potential concept retrieval stability in different networks and setups. For stability measurement, the number $N$ of CAV retrieval runs with different initialization parameters is set to $15$, which is similar to the ensemble size in~\cite{rabold_expressive_2020}, as we observed it is a good trade-off regarding computational speed. In each run, we utilize $100$ samples per concept, dividing them into $80$ for concept extraction and $20$ for validation (estimation of $f1$).

To further examine the influence of the number of concept training samples on CAV stability, we also test three additional setups with $20$, $40$, and $60$ training concept samples. The test has been conducted for all six networks.

\medskip\noindent\textbf{Experiment 2: Gradient Stability in Concept Detection.}
For gradient stability experiments, ResNet and YOLO5 are selected as models with the best CAV stability from Experiment 2. Moreover, we validate setups with 1D- and 3D-CAVs to see how gradient instability affects concept attribution in CAVs of different dimensionality. For the computation of SmoothGrad, we use the hyperparameter values recommended in~\cite{smilkov2017smoothgrad}: the number of noisy copies $N$ is set to 50, and the amount of applied Gaussian noise is set to 10\%.

\section{Experimental Results}
\label{sec:experiments}

\begin{table}

\setlength{\tabcolsep}{2pt}
\renewcommand{\arraystretch}{0.95}
\centering
\caption{Stability of generated CAVs of different dimensions for YOLO5.
}
\label{tab:cav-stab-yolo-v1}

\begin{tabular}{c|l|cccccccccc}

\hline
\multicolumn{2}{c|}{CAV} & $l_1$ & $l_2$ & $l_3$ & $l_4$ & $l_5$ & $l_6$ & $l_7$ & $l_8$ & $l_9$ & $l_{10}$ \\
\hline

\multirow{3}{*}{$cos$} & $1D$ & \textbf{0.977} & \textbf{0.980} & \textbf{0.972} & \textbf{0.971} & \textbf{0.956} & \textbf{0.955} & \textbf{0.859} & \textbf{0.927} & \textbf{0.923} & \textbf{0.929} \\
 & $2D$ & 0.522 & 0.342 & 0.483 & 0.526 & 0.729 & 0.590 & 0.670 & 0.670 & 0.715 & 0.666 \\
 & $3D$ & 0.346 & 0.378 & 0.467 & 0.553 & 0.577 & 0.617 & 0.664 & 0.707 & 0.652 & 0.602 \\
\hline

\multirow{3}{*}{$f1$} & $1D$ & \textbf{0.749} & \textbf{0.763} & \textbf{0.854} & \textbf{0.904} & \textbf{0.930} & \textbf{0.958} & \textbf{0.956} & \textbf{0.924} & \textbf{0.909} & \textbf{0.906} \\
 & $2D$ & 0.427 & 0.404 & 0.400 & 0.458 & 0.499 & 0.488 & 0.547 & 0.571 & 0.558 & 0.523 \\
 & $3D$ & 0.576 & 0.592 & 0.663 & 0.723 & 0.858 & 0.872 & 0.941 & 0.884 & 0.876 & 0.852 \\
\hline

\multirow{3}{*}{$S_{L_k}$} & $1D$ & \textbf{0.732} & \textbf{0.748} & \textbf{0.830} & \textbf{0.878} & \textbf{0.889} & \textbf{0.915} & \textbf{0.821} & \textbf{0.857} & \textbf{0.839} & \textbf{0.841} \\
 & $2D$ & 0.223 & 0.138 & 0.193 & 0.241 & 0.364 & 0.288 & 0.366 & 0.383 & 0.399 & 0.349 \\
 & $3D$ & 0.199 & 0.224 & 0.310 & 0.400 & 0.495 & 0.538 & 0.625 & 0.626 & 0.571 & 0.513 \\
\hline

\end{tabular}
\end{table}

\begin{table}

\setlength{\tabcolsep}{2pt}
\renewcommand{\arraystretch}{0.95}
\centering
\caption{Stability of generated CAVs of different dimensions for RCNN. 
}
\label{tab:cav-stab-rcnn-v1}

\begin{tabular}{c|l|cccccccccc}

\hline
\multicolumn{2}{c|}{CAV} & $l_1$ & $l_2$ & $l_3$ & $l_4$ & $l_5$ & $l_6$ & $l_7$ & $l_8$ & $l_9$ & $l_{10}$ \\
\hline

\multirow{3}{*}{$cos$} & $1D$ & \textbf{0.965} & \textbf{0.977} & \textbf{0.979} & \textbf{0.976} & \textbf{0.970} & \textbf{0.973} & \textbf{0.980} & \textbf{0.946} & \textbf{0.933} & \textbf{0.893} \\
 & $2D$ & 0.243 & 0.565 & 0.539 & 0.349 & 0.480 & 0.612 & 0.672 & 0.514 & 0.684 & 0.832 \\
 & $3D$ & 0.271 & 0.649 & 0.436 & 0.393 & 0.626 & 0.650 & 0.605 & 0.577 & 0.638 & 0.688 \\
\hline

\multirow{3}{*}{$f1$} & $1D$ & 0.528 & \textbf{0.588} & \textbf{0.730} & 0.550 & \textbf{0.762} & \textbf{0.809} & \textbf{0.724} & \textbf{0.888} & \textbf{0.946} & \textbf{0.944} \\
 & $2D$ & 0.530 & 0.582 & 0.533 & 0.420 & 0.486 & 0.506 & 0.448 & 0.543 & 0.521 & 0.659 \\
 & $3D$ & \textbf{0.536} & 0.552 & 0.586 & \textbf{0.563} & 0.680 & 0.741 & 0.637 & 0.753 & 0.873 & 0.941 \\
\hline

\multirow{3}{*}{$S_{L_k}$} & $1D$ & \textbf{0.509} & \textbf{0.574} & \textbf{0.715} & \textbf{0.537} & \textbf{0.739} & \textbf{0.787} & \textbf{0.710} & \textbf{0.840} & \textbf{0.882} & \textbf{0.843} \\
 & $2D$ & 0.129 & 0.329 & 0.287 & 0.147 & 0.233 & 0.309 & 0.301 & 0.279 & 0.357 & 0.548 \\
 & $3D$ & 0.145 & 0.358 & 0.255 & 0.221 & 0.426 & 0.482 & 0.385 & 0.435 & 0.557 & 0.647 \\
\hline

\end{tabular}
\end{table}

\begin{table}

\setlength{\tabcolsep}{2pt}
\renewcommand{\arraystretch}{0.95}
\centering
\caption{Stability of generated CAVs of different dimensions for SSD.
}
\label{tab:cav-stab-ssd-v1}

\begin{tabular}{c|l|cccccccccc}

\hline
\multicolumn{2}{c|}{CAV} & $l_1$ & $l_2$ & $l_3$ & $l_4$ & $l_5$ & $l_6$ & $l_7$ & $l_8$ & $l_9$ & $l_{10}$ \\
\hline

\multirow{3}{*}{$cos$} & $1D$ & \textbf{0.972} & \textbf{0.965} & \textbf{0.961} & \textbf{0.947} & \textbf{0.954} & \textbf{0.945} & \textbf{0.916} & \textbf{0.917} & \textbf{0.927} & \textbf{0.933} \\
 & $2D$ & 0.549 & 0.574 & 0.670 & 0.672 & 0.694 & 0.730 & 0.801 & 0.809 & 0.868 & 0.923 \\
 & $3D$ & 0.244 & 0.370 & 0.504 & 0.547 & 0.549 & 0.580 & 0.777 & 0.789 & 0.789 & 0.844 \\
\hline

\multirow{3}{*}{$f1$} & $1D$ & \textbf{0.666} & \textbf{0.758} & \textbf{0.909} & \textbf{0.888} & \textbf{0.949} & \textbf{0.962} & \textbf{0.897} & 0.846 & \textbf{0.874} & \textbf{0.858} \\
 & $2D$ & 0.413 & 0.418 & 0.440 & 0.406 & 0.429 & 0.421 & 0.492 & 0.523 & 0.652 & 0.614 \\
 & $3D$ & 0.556 & 0.596 & 0.636 & 0.738 & 0.831 & 0.891 & 0.870 & \textbf{0.880} & 0.856 & 0.843 \\
\hline

\multirow{3}{*}{$S_{L_k}$} & $1D$ & \textbf{0.647} & \textbf{0.731} & \textbf{0.873} & \textbf{0.841} & \textbf{0.905} & \textbf{0.909} & \textbf{0.821} & \textbf{0.776} & \textbf{0.810} & \textbf{0.801} \\
 & $2D$ & 0.227 & 0.240 & 0.295 & 0.273 & 0.298 & 0.307 & 0.394 & 0.423 & 0.566 & 0.567 \\
 & $3D$ & 0.136 & 0.221 & 0.320 & 0.404 & 0.456 & 0.517 & 0.676 & 0.694 & 0.675 & 0.712 \\
\hline

\end{tabular}
\end{table}

\begin{table}

\setlength{\tabcolsep}{2pt}
\renewcommand{\arraystretch}{0.95}
\centering
\caption{Stability of generated CAVs of different dimensions for ResNet. 
}
\label{tab:cav-stab-resnet-v1}

\begin{tabular}{c|l|ccccccc}

\hline
\multicolumn{2}{c|}{CAV} & $l_1$ & $l_2$ & $l_3$ & $l_4$ & $l_5$ & $l_6$ & $l_7$ \\
\hline

\multirow{3}{*}{$cos$} & $1D$ & \textbf{0.969} & \textbf{0.955} & \textbf{0.959} & \textbf{0.919} & 0.953 & 0.882 & 0.869 \\
 & $2D$ & 0.861 & 0.918 & 0.839 & 0.906 & \textbf{0.972} & \textbf{0.945} & \textbf{0.976} \\
 & $3D$ & 0.726 & 0.684 & 0.672 & 0.749 & 0.705 & 0.624 & 0.648 \\
\hline

\multirow{3}{*}{$f1$} & $1D$ & 0.668 & 0.856 & 0.847 & 0.910 & 0.944 & 0.983 & 0.960 \\
 & $2D$ & 0.402 & 0.369 & 0.406 & 0.588 & 0.598 & 0.356 & 0.423 \\
 & $3D$ & \textbf{0.716} & \textbf{0.867} & \textbf{0.871} & \textbf{0.920} & \textbf{0.956} & \textbf{0.988} & \textbf{0.967} \\
\hline

\multirow{3}{*}{$S_{L_k}$} & $1D$ & \textbf{0.647} & \textbf{0.817} & \textbf{0.812} & \textbf{0.836} & \textbf{0.900} & \textbf{0.868} & \textbf{0.834} \\
 & $2D$ & 0.346 & 0.339 & 0.340 & 0.533 & 0.581 & 0.336 & 0.413 \\
 & $3D$ & 0.520 & 0.593 & 0.585 & 0.689 & 0.673 & 0.616 & 0.626 \\
\hline

\end{tabular}
\end{table}

\begin{table}

\setlength{\tabcolsep}{2pt}
\renewcommand{\arraystretch}{0.95}
\centering
\caption{Stability of generated CAVs of different dimensions for SqueezeNet. 
}
\label{tab:cav-stab-sn-v1}

\begin{tabular}{c|l|ccccccc}

\hline
\multicolumn{2}{c|}{CAV} & $l_1$ & $l_2$ & $l_3$ & $l_4$ & $l_5$ & $l_6$ & $l_7$ \\
\hline

\multirow{3}{*}{$cos$} & $1D$ & \textbf{0.935} & \textbf{0.959} & \textbf{0.935} & \textbf{0.922} & \textbf{0.918} & \textbf{0.924} & \textbf{0.905} \\
 & $2D$ & 0.806 & 0.839 & 0.799 & 0.862 & 0.847 & 0.860 & 0.809 \\
 & $3D$ & 0.773 & 0.750 & 0.779 & 0.758 & 0.795 & 0.807 & 0.760 \\
\hline

\multirow{3}{*}{$f1$} & $1D$ & 0.547 & 0.654 & 0.863 & \textbf{0.883} & \textbf{0.920} & \textbf{0.948} & \textbf{0.968} \\
 & $2D$ & 0.381 & 0.364 & 0.377 & 0.409 & 0.453 & 0.551 & 0.506 \\
 & $3D$ & \textbf{0.620} & \textbf{0.668} & \textbf{0.863} & 0.877 & 0.911 & 0.932 & 0.961 \\
\hline

\multirow{3}{*}{$S_{L_k}$} & $1D$ & \textbf{0.511} & \textbf{0.627} & \textbf{0.807} & \textbf{0.815} & \textbf{0.845} & \textbf{0.876} & \textbf{0.876} \\
 & $2D$ & 0.307 & 0.306 & 0.301 & 0.352 & 0.384 & 0.474 & 0.409 \\
 & $3D$ & 0.479 & 0.501 & 0.673 & 0.665 & 0.724 & 0.752 & 0.731 \\
\hline

\end{tabular}
\end{table}

\begin{table}

\setlength{\tabcolsep}{2pt}
\renewcommand{\arraystretch}{0.95}
\centering
\caption{Stability of generated CAVs of different dimensions for EfficientNet.
}
\label{tab:cav-stab-en-v1}

\begin{tabular}{c|l|ccccccc}

\hline
\multicolumn{2}{c|}{CAV} & $l_{1}$ & $l_{2}$ & $l_{3}$ & $l_{4}$ & $l_{5}$ & $l_{6}$ & $l_{7}$ \\
\hline

\multirow{3}{*}{$cos$} & $1D$ & \textbf{0.924} & \textbf{0.929} & \textbf{0.936} & \textbf{0.933} & \textbf{0.892} & \textbf{0.898} & 0.767 \\
 & $2D$ & 0.773 & 0.751 & 0.711 & 0.772 & 0.754 & 0.769 & \textbf{0.884} \\
 & $3D$ & 0.787 & 0.744 & 0.770 & 0.835 & 0.668 & 0.843 & 0.638 \\
\hline

\multirow{3}{*}{$f1$} & $1D$ & \textbf{0.377} & 0.628 & 0.810 & \textbf{0.922} & 0.954 & \textbf{0.986} & 0.978 \\
 & $2D$ & 0.337 & 0.506 & 0.483 & 0.526 & 0.540 & 0.561 & 0.580 \\
 & $3D$ & 0.370 & \textbf{0.688} & \textbf{0.836} & \textbf{0.922} & \textbf{0.960} & 0.968 & \textbf{0.979} \\
\hline

\multirow{3}{*}{$S_{L_k}$} & $1D$ & \textbf{0.348} & \textbf{0.583} & \textbf{0.758} & \textbf{0.860} & \textbf{0.851} & \textbf{0.885} & \textbf{0.750} \\
 & $2D$ & 0.260 & 0.380 & 0.344 & 0.406 & 0.407 & 0.431 & 0.513 \\
 & $3D$ & 0.291 & 0.512 & 0.643 & 0.770 & 0.641 & 0.816 & 0.625 \\
\hline

\end{tabular}
\end{table}

\begin{figure*}[t]
  \centering
  \includegraphics[width=\linewidth]{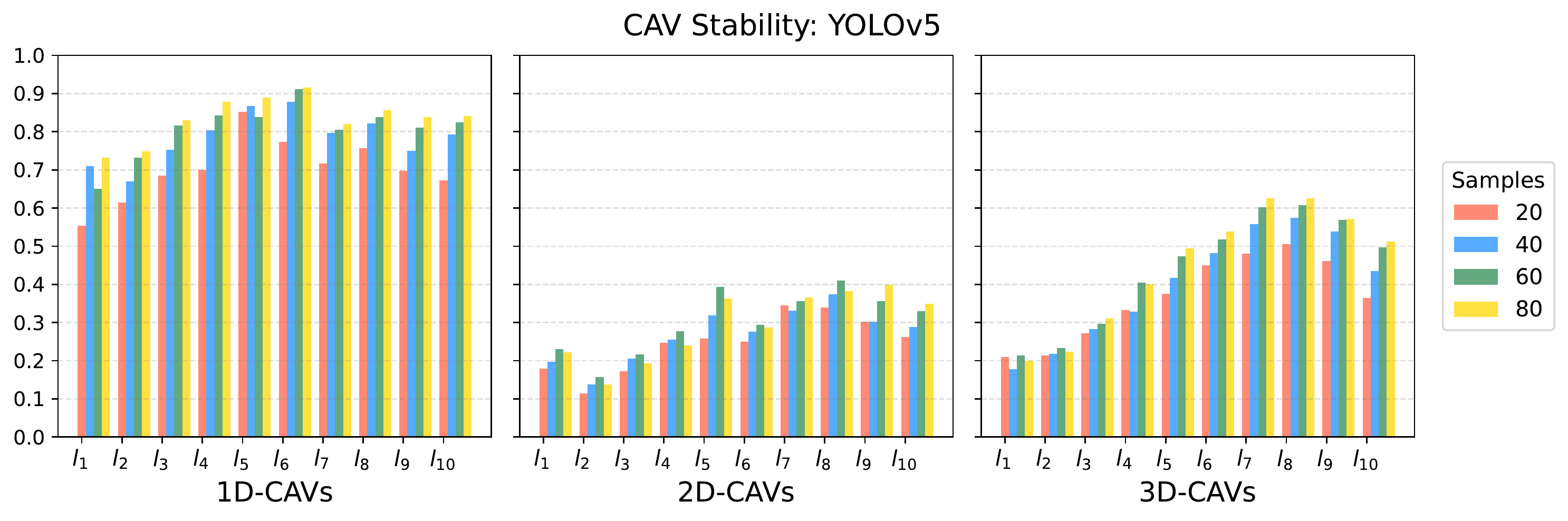}
  \caption{Impact of number of concept samples on CAVs stability for YOLO5}
  \label{fig:cav-n-samples-yolo}
\end{figure*}

\begin{figure*}[t]
  \centering
  \includegraphics[width=\linewidth]{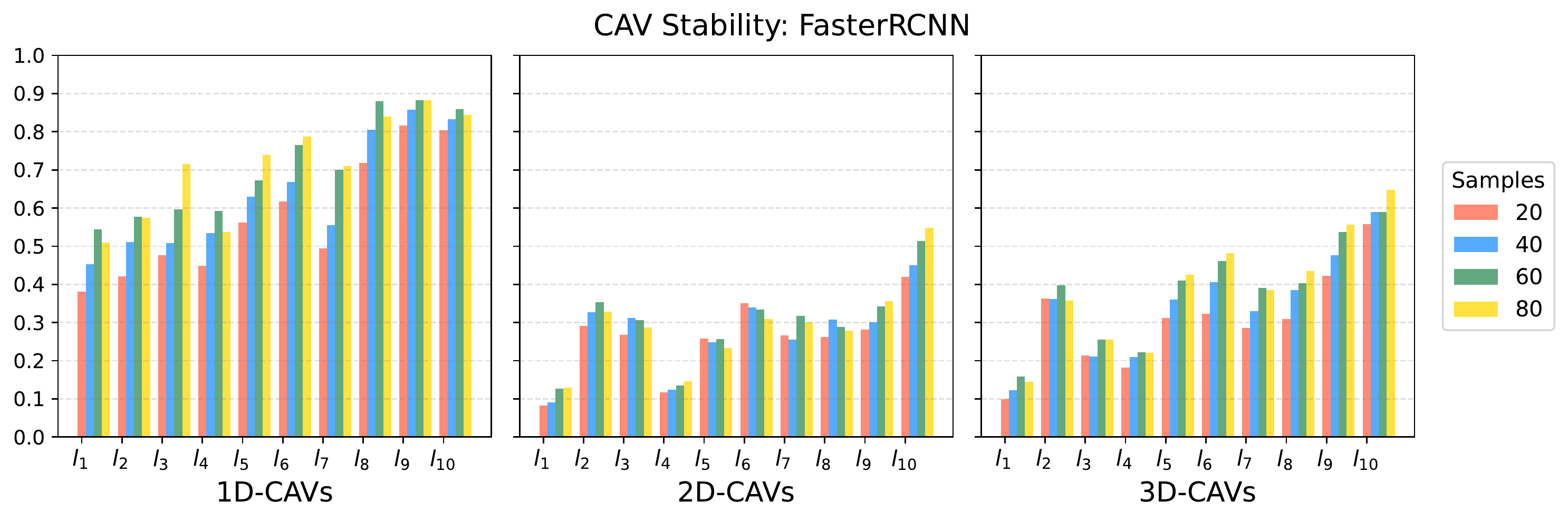}
  \caption{Impact of number of concept samples on CAVs stability for RCNN}
  \label{fig:cav-n-samples-rcnn}
\end{figure*}

\begin{figure*}[t]
  \centering
  \includegraphics[width=\linewidth]{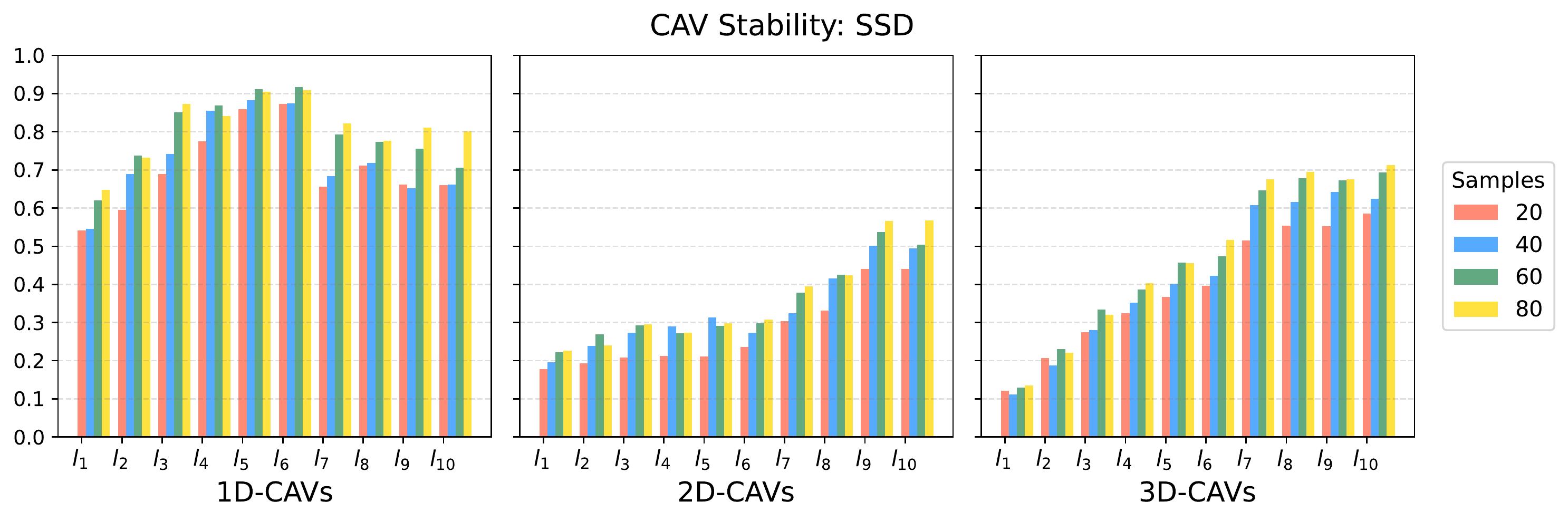}
  \caption{Impact of number of concept samples on CAVs stability for SSD}
  \label{fig:cav-n-samples-ssd}
\end{figure*}

\begin{figure*}[t]
  \centering
  \includegraphics[width=\linewidth]{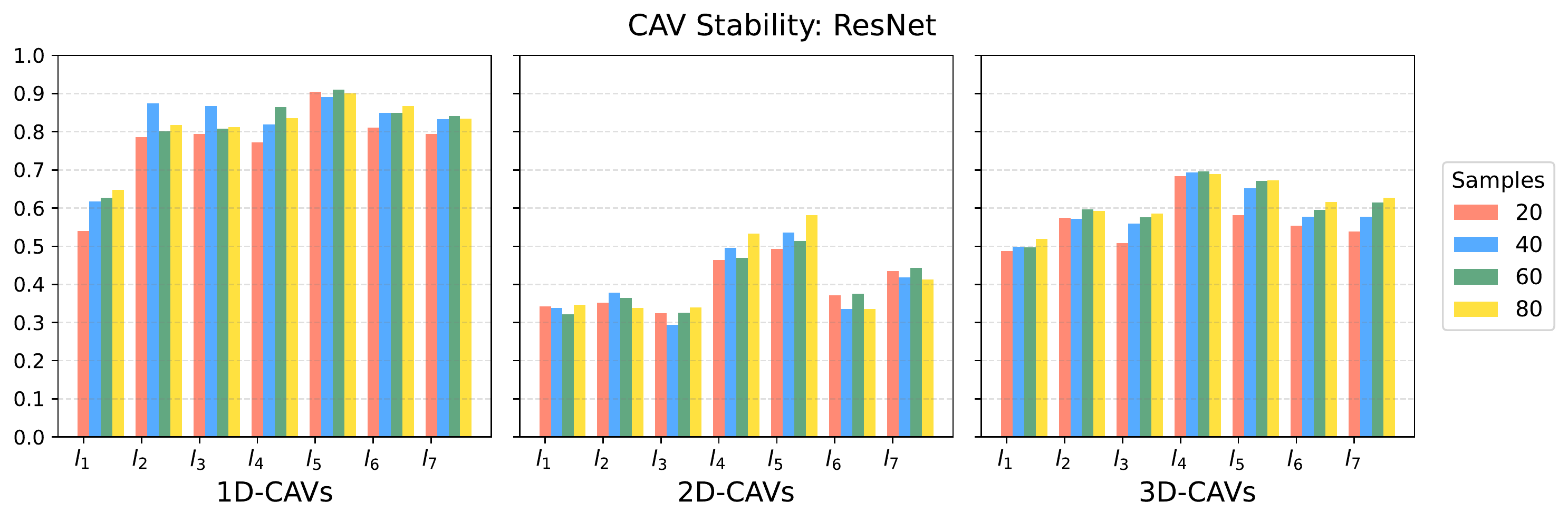}
  \caption{Impact of number of concept samples on CAVs stability for ResNet}
  \label{fig:cav-n-samples-resnet}
\end{figure*}

\begin{figure*}[t]
  \centering
  \includegraphics[width=\linewidth]{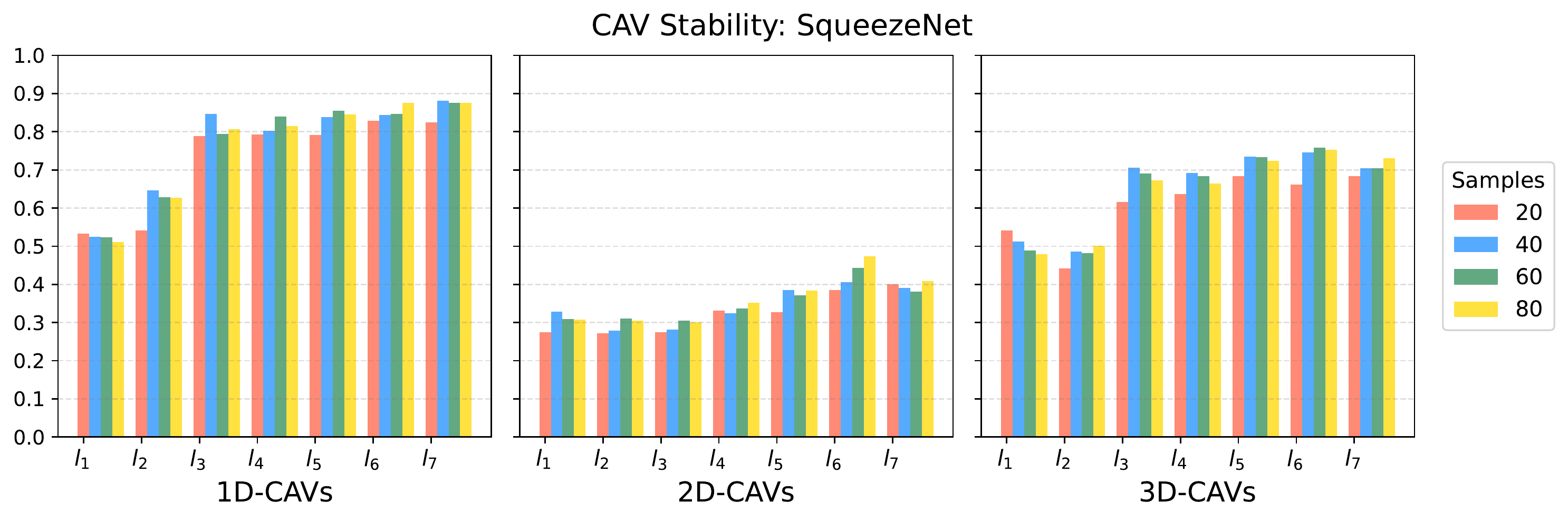}
  \caption{Impact of number of concept samples on CAVs stability for SqueezeNet}
  \label{fig:cav-n-samples-sn}
\end{figure*}

\begin{figure*}[t]
  \centering
  \includegraphics[width=\linewidth]{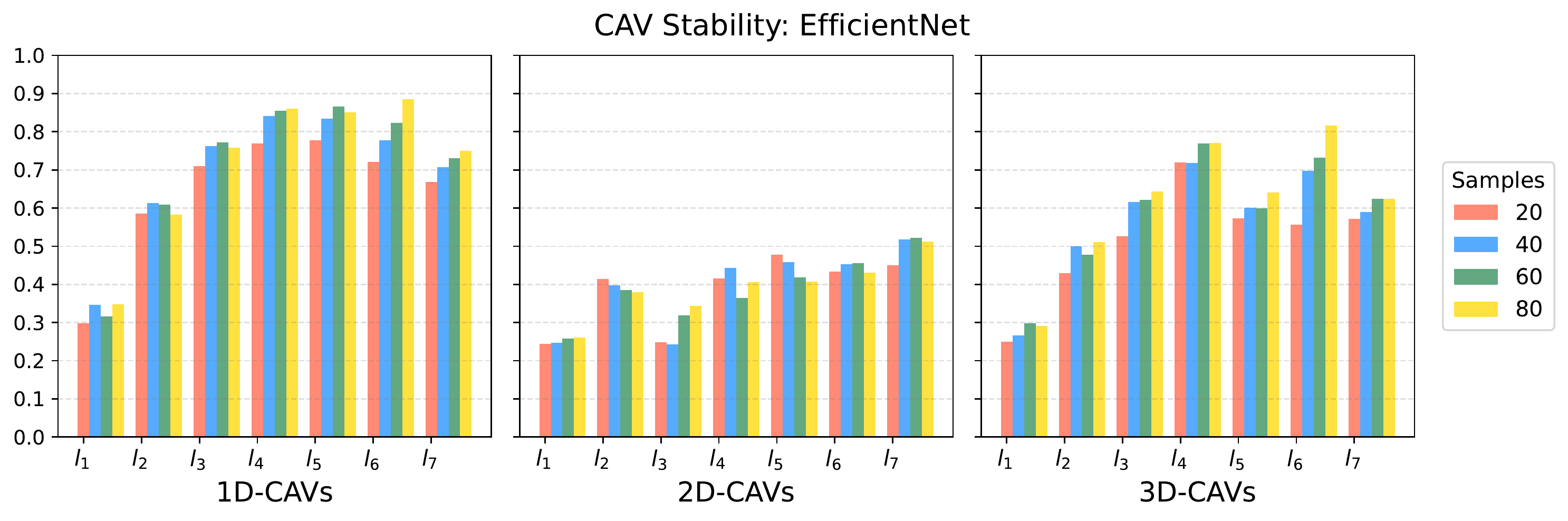}
  \caption{Impact of number of concept samples on CAVs stability for EfficientNet}
  \label{fig:cav-n-samples-en}
\end{figure*}

\subsection{CAV Stability and Dimensionality}
\label{sec:experiments-cav-stability}

The CAV stability results for 1D-, 2D- and 3D-CAVs in different layers of  YOLO5, RCNN, SSD,  ResNet, SqueezeNet, and EfficientNet networks are presented in Tabs.\,\ref{tab:cav-stab-yolo-v1}
to \ref{tab:cav-stab-en-v1}. In addition, Figs.\,\ref{fig:cav-n-samples-yolo} to
\ref{fig:cav-n-samples-en} visualize the impact of number of training concept samples on the overall stability of 1D-, 2D- and 3D-CAVs.

\medskip\noindent\textbf{CAV Dimensionality Impact.}
3D-CAVs are obtained without intermediate representation aggregation, and they demonstrate good concept separation ($f1$) that can sometimes even outperform that of 1D-CAVs. This is typical for classifiers, where, for instance, in all layers of ResNet (Tab.\,\ref{tab:cav-stab-resnet-v1})  $f1$ of 3D-CAVs is the highest. However, they for all models exhibit mediocre CAV consistency ($cos$), possibly due to the larger number of parameters and a relatively small number of training concept samples. Overall, 3D-CAVs are less stable than 1D-CAVs, but still can be used for CA.

In contrast, 2D-CAVs exhibit relatively high consistency (e.g., in Tab.\,\ref{tab:cav-stab-resnet-v1}, layers $l_5$, $l_6$, and $l_7$ have the top $cos$ values for 2D-CAVs), but they have the worst concept separation ($f1$), as observed in all tables. As a result, the overall 2D-CAV stability in all models is the worst. In 2D-CAVs, no distinction is made between different channels in the latent space due to 3D-to-2D aggregation. The noticeable reduction of concept separation ($f1$) in 2D-CAVs reinforces the assumption made in other works (e.g.,~\cite{bau2017network,fong2018net2vec}) that concept information is encoded in different convolutional filters or their linear combinations.

1D-CAVs achieve the best overall CAV-stability due to their (mostly) best consistency ($cos$) and good concept separation ($f1$). Moreover, 1D-CAVs have the advantage of fast computation speed since they have fewer parameters. These unique features of 1D-CAVs make them highly stable even in shallow layers, where other CAVs may experience low stability. For example, in Tab.\,\ref{tab:cav-stab-yolo-v1}, the stability of 1D-CAVs in layer $l_1$ $S_{L_k}=0.732$ is substantially higher than that of 2D- and 3D-CAVs, which are only $0.223$ and $0.199$, respectively.

Based on our empirical findings, we recommend using 1D-CAV as the default representation for most applications due to its superior overall stability. However, for safety-critical applications, we advise using our stability assessment methodology prior to CA.

\medskip\noindent\textbf{Concept Abstraction Level Impact.}
In OD models, experiments are conducted with concepts of medium-to-high levels of abstraction (complex shapes and human body parts), which are usually detected in middle and deep layers of the network~\cite{wang2020chain}. Thus, it is expected that there will be worse concept separation ($f1$) in shallow layers, and this has indeed been observed across all dimension sizes of CAVs (as shown in Tabs.\,\ref{tab:cav-stab-yolo-v1}-\ref{tab:cav-stab-en-v1}).

However, this observation is not always valid for 2D-CAVs, as results have shown that concept separation drops in some deeper layers. For instance, in Tab.\,\ref{tab:cav-stab-rcnn-v1} $l_4$ and $l_7$ have $f1$ values $0.420$ and $0.448$, while for $l_1$ it is $0.530$. Also, Tab.\,\ref{tab:cav-stab-rcnn-v1} shows that the increase of $f1$ for 2D-CAVs is not as high as it is for 1D- and 3D-CAVs. The range of $f1$ for 2D-CAVs is between $0.420$ to $0.659$, whereas for 3D-CAVs, it is between $0.536$ to $0.941$. These findings further support the hypothesis that concept information is encoded in linear combinations of convolutional filters~\cite{bau2017network,fong2018net2vec}.

\medskip\noindent\textbf{Impact of Number of CAV Training Samples.}
Figs.\,\ref{fig:cav-n-samples-yolo} to \ref{fig:cav-n-samples-en} demonstrate that increasing the number of training concept images has a positive impact on the stability of CAV. However, labeling concepts is a time-consuming and expensive process. Therefore, we recommend using at least 40 to 60 concept-related samples for training each CAV. In most cases, the stability obtained with 80 samples is only marginally better than that obtained with 40 (see Fig.\,\ref{fig:cav-n-samples-sn}) or 60 samples (see Fig.\,\ref{fig:cav-n-samples-yolo} and Fig.\,\ref{fig:cav-n-samples-ssd}).

\medskip\noindent\textbf{CNN Architecture Impact.}
From Tabs.\,\ref{tab:cav-stab-yolo-v1} to
\ref{tab:cav-stab-en-v1} we see that top CAV stability ($S_{L_k}$) values achieved by ODs and classifiers for CAVs trained on the same concept datasets are very similar. However, due to architectural differences, the top stability values are achieved at different relative layer depths. For example, the top stabilities for 1D-CAVs in YOLO5, RCNN, and SSD object detectors are achieved in layers $l_6$, $l_9$, and $l_6$, respectively, with corresponding values of $0.915$, $0.882$, and $0.909$ (see Tabs.\,\ref{tab:cav-stab-yolo-v1}, \ref{tab:cav-stab-rcnn-v1}, and~\ref{tab:cav-stab-ssd-v1}). Similarly, the top stability values for 1D-CAVs for ResNet, SqueezeNet, and EfficientNet classifiers are achieved in layers $l_5$, $l_7$, and $l_6$, respectively, with corresponding values of $0.900$, $0.876$, and $0.885$ (Tab.\,\ref{tab:cav-stab-resnet-v1},~\ref{tab:cav-stab-sn-v1},~and~\ref{tab:cav-stab-en-v1}). The same tables show that the layers with top stability values may vary for different sizes of CAV dimensions even within the same model (e.g., in Tab.\,\ref{tab:cav-stab-yolo-v1}, the YOLO5 top stabilities for 1D-, 2D-, and 3D-CAV are obtained in layers $l_6$, $l_9$, and $l_8$, respectively).

The CAV stability differences among inspected architectures can also be observed in Figs.\,\ref{fig:cav-n-samples-yolo} to \ref{fig:cav-n-samples-en}. For example, in the case of 1D-CAV of ResNet (Fig.\,\ref{fig:cav-n-samples-resnet}) and 1D- and 3D-CAVs of SqueezeNet (Fig.\,\ref{fig:cav-n-samples-sn}), we observe that the stability value quickly reaches its optimal values in the first one or two layers and remains similar in deeper layers. In other cases, such as 3D-CAV of SSD (Fig.\,\ref{fig:cav-n-samples-ssd}) or all CAV dimensions of RCNN (Fig.\,\ref{fig:cav-n-samples-rcnn}), stability gradually increases with the relative depth of the layer. Finally, the stabilities of 1D- and 3D-CAVs of YOLO5 (Fig.\,\ref{fig:cav-n-samples-yolo}) or 1D- and 3D-CAVs of EfficientNet (Fig.\,\ref{fig:cav-n-samples-en}) grow until an optimal layer in the middle and slowly shrink after it.




\begin{table}

\setlength{\tabcolsep}{2pt}
\renewcommand{\arraystretch}{0.95}
\centering
\caption{Gradient stability in layers of ResNet for 1D-CAV.
}
\label{tab:grad-stability-rn1d-1d}

\begin{tabular}{c|ccccccc}

\hline
\multicolumn{1}{l|}{Measure} & $l_{1}$ & $l_{2}$ & $l_{3}$ & $l_{4}$ & $l_{5}$ & $l_{6}$ & $l_{7}$ \\

\hline
$TP$ & 669 & 748 & 731 & 709 & 728 & 677 & 500 \\
$TN$ & 718 & 697 & 704 & 727 & 724 & 791 & 1000 \\
\hline
$FP$ & 52 & 26 & 33 & 30 & 26 & 19 & 0 \\
$FN$ & 61 & 29 & 32 & 34 & 22 & 13 & 0 \\
\hline
Acc & 0.92 & 0.96 & 0.96 & 0.96 & 0.97 & 0.98 & 1.00 \\
\hline
CAD, $\%$ & 20.7 & 13.6 & 14.1 & 14.4 & 10.5 & 6.0 & 0.6 \\

\hline
\end{tabular}
\end{table}

\begin{table}

\setlength{\tabcolsep}{2pt}
\renewcommand{\arraystretch}{0.95}
\centering
\caption{Gradient stability in layers of ResNet for 3D-CAV.
}
\label{tab:grad-stability-rn3d-3d}

\begin{tabular}{c|ccccccc}

\hline
\multicolumn{1}{l|}{Measure} & $l_{1}$ & $l_{2}$ & $l_{3}$ & $l_{4}$ & $l_{5}$ & $l_{6}$ & $l_{7}$ \\

\hline
$TP$ & 667 & 709 & 693 & 716 & 886 & 789 & 500 \\
$TN$ & 687 & 721 & 721 & 709 & 561 & 688 & 1000 \\
\hline
$FP$ & 80 & 41 & 43 & 41 & 26 & 13 & 0 \\
$FN$ & 66 & 29 & 43 & 34 & 27 & 10 & 0 \\
\hline
Acc & 0.90 & 0.95 & 0.94 & 0.95 & 0.96 & 0.98 & 1.00 \\
\hline
CAD, $\%$ & 31.3 & 17.5 & 19.5 & 18.0 & 12.8 & 5.8 & 0.7 \\

\hline
\end{tabular}
\end{table}

\begin{table}

\setlength{\tabcolsep}{2pt}
\renewcommand{\arraystretch}{0.95}
\centering
\caption{Gradient stability in layers of YOLO5 for 1D-CAV.
}
\label{tab:grad-stability-yolo1d-1d}

\begin{tabular}{c|cccccccccc}

\hline
\multicolumn{1}{l|}{Measure} & $l_{1}$ & $l_{2}$ & $l_{3}$ & $l_{4}$ & $l_{5}$ & $l_{6}$ & $l_{7}$ & $l_{8}$ & $l_{9}$ & $l_{10}$ \\

\hline
$TP$ & 973 & 997 & 979 & 1012 & 998 & 989 & 998 & 1046 & 1065 & 771 \\
$TN$ & 1014 & 994 & 1009 & 990 & 989 & 1038 & 1069 & 1042 & 1024 & 1331 \\
\hline
$FP$ & 73 & 66 & 61 & 58 & 80 & 53 & 36 & 23 & 22 & 21 \\
$FN$ & 76 & 79 & 87 & 76 & 69 & 56 & 33 & 25 & 25 & 13 \\
\hline
Acc & 0.93 & 0.93 & 0.93 & 0.94 & 0.93 & 0.95 & 0.97 & 0.98 & 0.98 & 0.98 \\
\hline
CAD, $\%$ & 24.1 & 27.4 & 26.4 & 28.0 & 27.9 & 22.5 & 17.6 & 13.8 & 10.9 & 14.5 \\

\hline
\end{tabular}
\end{table}

\begin{table}

\setlength{\tabcolsep}{2pt}
\renewcommand{\arraystretch}{0.95}
\centering
\caption{Gradient stability in layers of YOLO5 for 3D-CAV.
}
\label{tab:grad-stability-yolo3d-3d}

\begin{tabular}{c|cccccccccc}

\hline
\multicolumn{1}{l|}{Measure} & $l_1$ & $l_2$ & $l_3$ & $l_4$ & $l_5$ & $l_6$ & $l_7$ & $l_8$ & $l_9$ & $l_{10}$ \\

\hline
$TP$ & 959 & 1010 & 985 & 985 & 973 & 1001 & 1054 & 1032 & 1039 & 785 \\
$TN$ & 1025 & 1012 & 1002 & 1007 & 980 & 990 & 1015 & 1057 & 1047 & 1321 \\
\hline
$FP$ & 75 & 58 & 74 & 74 & 87 & 80 & 28 & 22 & 29 & 11 \\
$FN$ & 77 & 56 & 75 & 70 & 96 & 65 & 39 & 25 & 21 & 19 \\
\hline
Acc & 0.93 & 0.95 & 0.93 & 0.93 & 0.91 & 0.93 & 0.97 & 0.98 & 0.98 & 0.99 \\
\hline
CAD, $\%$ & 27.2 & 29.6 & 28.5 & 29.7 & 30.6 & 29.2 & 15.7 & 14.3 & 11.5 & 14.1 \\

\hline
\end{tabular}
\end{table}

\subsection{Gradient Stability in Concept Detection}
\label{sec:experiments-gradient-stability}

Based on the experimental results, it can be concluded that the negative impact of gradient instability on concept analysis using TCAV is minimal. The results presented in Tabs.\,\ref{tab:grad-stability-rn1d-1d}~and~\ref{tab:grad-stability-rn3d-3d} are based on $1500$ concept attribution predictions (see Eq.\,\ref{eq:grad-stability-attr}) for $500$ images and $3$ concepts per image, for each tested layer of ResNet with 1D- and 3D-CAVs, respectively. Similarly, Tabs.\,\ref{tab:grad-stability-yolo1d-1d}~and~\ref{tab:grad-stability-yolo3d-3d} are built for each tested layer of YOLO5 with 1D- and 3D-CAVs, respectively, using $2136$ concept attribution predictions for $712$ bounding boxes and $3$ concepts per bounding box.

\medskip\noindent\textbf{SmoothGrad Impact.}
In the Tabs.\,\ref{tab:grad-stability-rn1d-1d} to \ref{tab:grad-stability-yolo3d-3d}, the relative depth of CNN backbone layers is increasing from left to right, while gradient backpropagation depth from outputs to CAV layer is increasing in right to left order. As expected, the gradient is becoming more unstable with backpropagation depth~\cite{smilkov2017smoothgrad}, resulting in higher $\text{CAD}$ values in shallow layers compared to deeper layers. 
The higher number of concept attribution sign flips is observed in shallow layers (see Sec.\,\ref{sec:method-metrics}), where accuracy (Acc) values in those layers are low. These observations confirm the negative correlation between CAD and Acc, where CAD increases as Acc decreases. This suggests that gradient smoothing techniques, such as SmoothGrad, can have a higher impact on concept attribution values in shallow layers, where the gradient instability is higher.

Despite the negative correlation between CAD and Acc values, the overall accuracy values remain above $0.9$ for all layers in the provided tables. The lowest accuracy value for ResNet of $\text{Acc}=0.90$ is observed in Tab.\,\ref{tab:grad-stability-rn3d-3d} for $l_1$. For YOLO5 the lowest $\text{Acc}=0.91$ is obtrained in $l_5$ (Tab.\,\ref{tab:grad-stability-yolo3d-3d}). This indicates that the sign of concept attribution is only changed for a minority of predictions across all tested networks and configurations. However, it is worth noting that CAD values can be high in shallow layers, for instance, $\text{CAD}=31.3\%$ at layer $l_1$ of Tab.\,\ref{tab:grad-stability-rn3d-3d}, resulting in a higher rate of concept attribution sign flipping compared to deeper layers.

The use of SmoothGrad comes at a higher computational cost compared to vanilla gradient. It is more than $N$ times (number of noisy copies) computationally expensive, and mostly impacts concept attribution in shallow and middle layers of networks. Therefore, it is advisable to use SmoothGrad when conducting concept analysis in shallow layers of networks with large backbones such as ResNet101 or ResNet152.

\medskip\noindent\textbf{CAV Dimensionality Impact.}
The use of 1D-CAV representations generally results in lower CAD values than 3D-CAVs, typically with a difference of 2-3\%. This behavior can be attributed to the higher stability of 1D-CAVs, which is in turn caused by the lower number of parameters. The observation is consistent across all layers of ResNet and the majority of YOLO5 layers, as shown in Tabs.\,\ref{tab:grad-stability-rn1d-1d} to \ref{tab:grad-stability-yolo3d-3d}. However, the dimensionality of CAV does not affect the behavior of gradient instability in other regards: CAD remains higher and Acc lower in shallow layers regardless of the CAV dimensionality.

\section{Conclusion and Outlook}
\label{sec:conclusion}

This study proposes a framework and metrics for evaluating the layer-wise stability of global vector representations in object detection and classification CNN models for explainability purposes. We introduced two stability metrics: concept retrieval stability and concept attribution stability. Also, we proposed adaptation methodologies for unsupervised CA and supervised gradient-based CA methods for combined, labeling-efficient application in object detection models. 

Our concept retrieval stability metric jointly evaluates the consistency and separation in the feature space of concept semantic concept representations obtained across multiple runs with different initialization parameters. We used the TCAV method as an example to examine factors that affect stability and found that aggregated 1D-CAV representations offer the best performance. Furthermore, we determined that a minimum of 60 training samples per concept is necessary to ensure high stability in most cases.

The second metric, concept attribution stability, assesses the impact of gradient smoothing techniques on the stability of concept attribution. Our observations suggest that 1D-CAVs are more resistant to gradient instability, particularly in deep layers, and we recommend using gradient smoothing in shallow layers of deep network backbones.

Our work provides valuable quantitative insights into the robustness of concept representation, which can inform the selection of network layers and concept representations for CA in safety-critical applications. For future work, it will be interesting to apply the proposed approaches and metrics to alternative global concept vector representations and perform comparative analysis.

\subsection*{Acknowledgments}

The research leading to these results is funded by the German Federal Ministry for Economic Affairs and Climate Action within the project “KI Wissen – Entwicklung von Methoden für die Einbindung von Wissen in maschinelles Lernen”. The authors would like to thank the consortium for the successful cooperation.

\bibliographystyle{splncs04}
\bibliography{ref}
\end{document}